\documentclass[letterpaper, 10 pt, conference]{ieeeconf} 

\IEEEoverridecommandlockouts      

\usepackage{cite}
\usepackage{amsmath,amssymb,amsfonts}
\usepackage{algorithmic}
\usepackage{graphicx}
\usepackage{textcomp}
\usepackage{mathtools}
\usepackage{bm}  
\usepackage{xcolor}
\usepackage{todonotes}
\usepackage{comment}
\usepackage{soul}
\usepackage{url}
\def\BibTeX{{\rm B\kern-.05em{\sc i\kern-.025em b}\kern-.08em
    T\kern-.1667em\lower.7ex\hbox{E}\kern-.125emX}}

\newcommand{\sub}[1]{\mathrm{#1}}

\usepackage{booktabs}
\newcommand{\ra}[1]{\renewcommand{\arraystretch}{#1}}
\usepackage{arydshln}
\usepackage{multirow}

\usepackage{algorithm}

\usepackage{subcaption}

\usepackage{pifont}
\usepackage{units}

\usepackage{dblfloatfix}

\newcommand{\vsigma}{\mbox{\boldmath $\sigma$}}
\newcommand{\vtau}{\mbox{\boldmath $\tau$}}

\newcommand{\va}{\bm a}

\newcommand{\vg}{\bm g}

\newcommand{\vm}{\bm m}
\newcommand{\vn}{\bm n}

\newcommand{\vp}{\bm p}
\newcommand{\vq}{\bm q}

\newcommand{\vv}{\bm v}

\newcommand{\vC}{\bm C}

\newcommand{\vF}{\bm F}

\newcommand{\vI}{\bm I}
\newcommand{\vJ}{\bm J}
\newcommand{\vK}{\bm K}

\newcommand{\vM}{\bm M}

\newcommand{\vS}{\bm S}

\begin{document}

\title{\LARGE \bf
Collision detection and identification for a legged manipulator
}

\author{Jessie van Dam, Andreea Tulbure, Maria Vittoria Minniti, Firas Abi-Farraj, Marco Hutter 
\thanks{This work was supported in part by the Swiss National Science Foundation through the National Centre of Competence in Research Robotics (NCCR Robotics), in part by the Swiss National Science Foundation through the National Centre of Competence in Digital Fabrication (NCCR dfab), in part by armasuisse and in part by the Swiss National Science Foundation (SNSF) as part of project No.188596.} 
\thanks{All authors are with the Robotic Systems Lab, ETH Zurich, Zurich 8092, Switzerland.}%
}

\maketitle
\pagestyle{empty}

\begin{abstract}
To safely deploy legged robots in the real world it is necessary to provide them with the ability to reliably detect unexpected contacts and accurately estimate the corresponding contact force. 
In this paper, we propose a collision detection and identification pipeline for a quadrupedal manipulator. We first introduce an approach to estimate the collision time span based on band-pass filtering and show that this information is key for obtaining accurate collision force estimates. We then improve the accuracy of the identified force magnitude by compensating for model inaccuracies, unmodeled loads, and any other potential source of quasi-static disturbances acting on the robot. We validate our framework with extensive hardware experiments in various scenarios, including trotting and additional unmodeled load on the robot. 

\end{abstract}

\section{Introduction} \label{sec:introduction}

Quadrupedal robots have recently become sufficiently advanced to be deployed in unknown and unstructured environments, where they could operate alongside humans or other robots. In such settings, unexpected collisions with people or objects are likely to occur and collision detection plays an important role in ensuring the safety of the external environment, and also for keeping the balance of the robot. Thus, robots need to be able to reliably detect such collisions, accurately estimate the corresponding contact forces and their location, and react accordingly.

Literature refers to the framework for responding to collisions as the \textit{collision event pipeline}, separating a collision event into five phases, i.e.: detection, isolation, identification, classification and reaction \cite{Haddadin2017RobotIdentification}. In this paper, we focus on two phases.
Briefly, the \textit{collision detection} phase defines when a collision happens based on the external estimated torques acting on the robot, and the \textit{collision identification} phase estimates the external collision force in magnitude and direction.

Momentum-based collision event pipelines have proven to be successful in collision handling for fixed-based manipulators \cite{Haddadin2017RobotIdentification, Mamedov2020PracticalDetection}. However, especially for complex high-DoF robots, the employed robot model might be affected by potential accidents and modeling inaccuracies due to, for example, static friction, wear and tear. In addition, it might be difficult to model all kinds of payloads that the robot might need to carry during a manipulation task.
Such elements of uncertainty can lead to failures of the collision detection phase (e.g., detecting a collision when there is none) or inaccurate external force estimation (e.g., considering the force coming from such disturbances as being part of the collision force). Therefore, in this work, we focus on the design of a collision detection and identification method that is robust against such disturbance factors. To achieve this, we recognize the importance of detecting the start and end time of a collision, and use this information to improve the collision force estimation during the identification phase. 
We provide a comparison of state-of-the-art methods focusing on torque estimation and collision detection. Finally, we carry out extensive hardware tests where we validate our pipeline and open-source the data of over 400 collisions from our experiments as a public benchmark dataset.

\begin{figure}[!t]
\centerline{\includegraphics[width=1.0\columnwidth ]{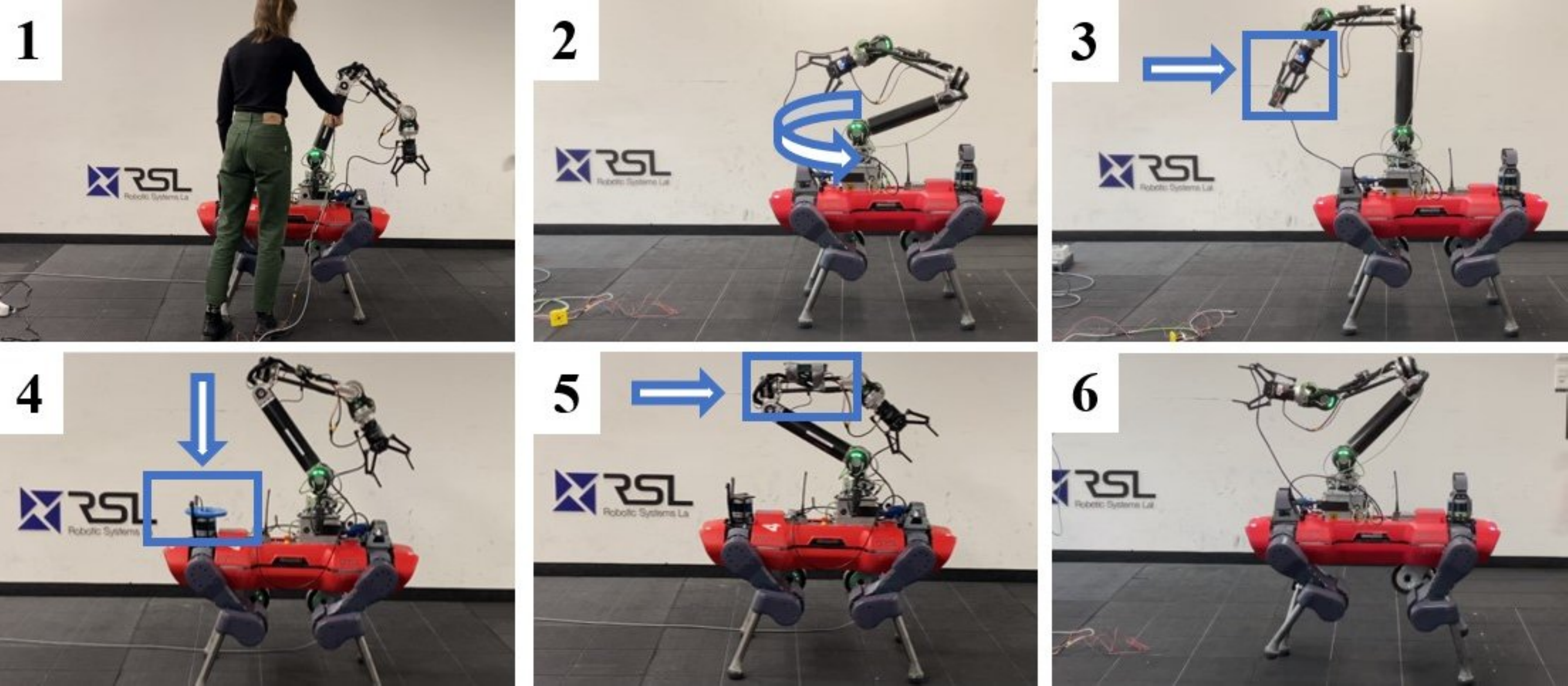}}
\caption{The quadrupedal robot ANYmal with a 6-DoF manipulator mounted on top. We consider the following scenarios: (1) stance; (2) arm motion; (3) measured load in the gripper; (4) unmodeled load on the base; (5) unmodeled load on the forearm; and (6) trot.}
\label{fig:experiments}
\end{figure}

\subsection{Related work}
\label{sec:related-work}

Existing collision event pipelines for robotic systems are based on monitoring the generalized external torques acting on the robot. Many torque estimation methods have been proposed in the literature, e.g. direct estimation \cite{Ito2019ExperimentalRobot}, static direct estimation \cite{Mattioli2017InteractionRobots}, filtered dynamics observer \cite{VanDamme2011EstimatingMeasurements} or momentum-based observers (MBO) \cite{Vorndamme2017CollisionHumanoids}, \cite{Bledt2018ContactTerrains}, \cite{Morlando2021Whole-bodyRobots}, \cite{Wahrburg2015CartesianMomentum}, \cite{Garofalo2019SlidingAcceleration}. 
For collision detection, the estimated torques are usually compared with appropriately-designed thresholds \cite{DeLuca2008ExploitingReaction, DeLuca2006CollisionArm}.  Constant thresholds suffer from the problem of \textit{false positives} (FPs) occurring when a robot task includes high-frequency content or \textit{false negatives} (FNs) for light impact collisions. 
Another challenge in collision detection is the presence of modeling errors that can generate slow and configuration-dependent variations in the estimated torques. Dynamic thresholds based on velocity \cite{Geravand2013Human-robotArchitecture, Qiu2020AdaptiveEstimation} or standard deviation of the estimated force \cite{Escobedo2021ContactSensors} have been proposed to cope with such a problem.
An effective way to improve the robustness of detection in the presence of model uncertainties \cite{Cho2013Adaptation-and-collisionInteraction} or unknown loads \cite{Cho2013CollisionUncertainty} is to filter the estimated torques with a \textit{band-pass filter} (BPF). BPF-based methods filter out the quasi-static component in the estimated external torques due to payloads or model inaccuracies, allowing to reduce the detection threshold by an order of five \cite{Cho2013Adaptation-and-collisionInteraction}, and thus obtain faster detection. However, so far BPF-based works have only considered high-impact contacts on low-DoF manipulators and have focused only on the detection phase. 

The collision detection step is usually followed by a collision isolation phase \cite{Faraji2016DesigningDynamics,Haddadin2017RobotIdentification,Flacco2016Residual-basedRobots} and a collision identification phase. Collision identification has been studied for fixed-base manipulators \cite{Haddadin2017RobotIdentification}, and works have also considered wheeled humanoids \cite{Bolotnikoval2019CompliantObserver, Bolotnikova2018ContactDiscrepancies}. Legged robots, on the other hand, impose additional challenges, such as the high number of DoFs which increases the computational time and tuning complexity~\cite{Faraji2016DesigningDynamics}, impulsive contact switching during trotting \cite{Morlando2021Whole-bodyRobots}, and the necessity of reasoning about a large number of contacts \cite{Vorndamme2017CollisionHumanoids}.
Thus, much interest has recently been given to tackling the problem of force estimation and collision identification for legged robots \cite{Mattioli2017InteractionRobots, Hawley2019ExternalRobots, Faraji2016DesigningDynamics, Morlando2021Whole-bodyRobots, Flacco2016Residual-basedRobots, Cong2020ContactControl, Vorndamme2017CollisionHumanoids}. 
In \cite{Mattioli2017InteractionRobots}, contacts are applied while a NAO humanoid is at standstill or static equilibrium. Similarly, in \cite{Ito2019ExperimentalRobot}, four different collisions between $\unit[25-30]{N}$ lasting $\unit[18]{s}$ are applied on a HRP-4 humanoid. Collision identification during walking with a NAO robot is addressed in \cite{Hawley2019ExternalRobots}; the authors estimate the external collision forces using internal (\textit{proprioceptive}) sensing in combination with an Inertial Measurement Unit (IMU) and force-sensing resistors beneath the feet.
Contact wrench estimation for quadrupeds has mostly been restricted to estimating the forces at the feet in contact with the ground \cite{Cong2020ContactControl, Yang2019SlipRobots, Bledt2018ContactTerrains}. 
In \cite{Morlando2021Whole-bodyRobots}, external forces are estimated at unknown contact points other than the feet; multiple case studies in which collisions are applied to the legs, both during stance and trotting, are validated in simulation. 

However, such works do not consider the effect of model uncertainties or unmodeled loads on the estimated external forces. In \cite{Vorndamme2017CollisionHumanoids}, force/torque (F/T) sensors placed on the robot body are used to compensate for the weight of an unknown load; model inaccuracies are not addressed and the method is only validated in simulation.
Furthermore, previous works on collision handling for legged robots have only been evaluated on real hardware on a small number of collisions. 

\subsection{Contributions}
In this work, we propose a model-based collision detection and identification framework for a quadrupedal manipulator, able to estimate the collision time span and external collision forces in various scenarios, including unmodeled loads on the robot.
The main contributions of this work are the following:
\begin{itemize}
    \item A BPF-based approach for estimating the time span of applied collisions, that is shown to make the collision identification phase more accurate.
    \item An improved identification method, based on continuous disturbance force estimation, which can compensate for unmodeled loads and model inaccuracies.
    \item A comparison of state-of-the-art external torque estimation and collision detection methods.
    \item Extensive experimental validation of the proposed framework on a quadrupedal manipulator. This is accompanied by an open-source dataset \footnote{Available at \url{https://u.ethz.ch/wmIqO}} of multiple collision samples, including collisions on the base and arm in various scenarios as depicted in Fig.~\ref{fig:experiments}.
\end{itemize}

\section{Preliminaries} \label{sec:preliminaries}

In this section,
we first present the model of a floating-base manipulator (Sec.~\ref{sec:floating_base_model}). Then, we give the fundamentals of the first block of the collision pipeline (see Fig.~\ref{fig:collision_pipeline}). These fundamentals are generally applicable to any robot system and include the estimation of the external generalized torques (Sec.~\ref{sec:torque_estimation}) and the external wrench (Sec.~\ref{sec:wrench_estimation}).

Throughout the paper, we use the symbol $\hat{a}$ to denote the estimate of a variable $a$. In addition, the 2-norm of a vector $\va$ is denoted as $|\va|$, while the absolute value of one of the components of the vector is denoted as $|a|$. 
\subsection{Floating-base dynamic model}
\label{sec:floating_base_model}
The equations of motion of a floating-base robot are given by \cite{Sleiman2021AManipulation,Flacco2016Residual-basedRobots}
\begin{equation} \label{eq:dynamic_model_floating_base}
\vM(\vq) \dot{\vv}+\vC(\vq, \vv)\vv + \vg(\vq)  =\vS^{T} \vtau_{\sub{m}} + \vtau_{\sub{ext}} + \vtau_{\sub{ft}} ,
\end{equation}
where $\vq, \vv \in \mathbb{R}^{6+n}$ are the robot generalized coordinates and velocities, respectively, and $n$ is the number of actuated joints. $\vM(\vq)\in \mathbb{R}^{(6+n)\times (6+n)}$ 
is the inertia matrix, 
$\vC(\vq, \vv) \in \mathbb{R}^{(6+n)\times (6+n)}$ 
is the Coriolis matrix while $\boldsymbol{g}(\boldsymbol{q})\in \mathbb{R}^{6+n}$ 
is the vector of gravitational terms. The joint torques $\vtau_\sub{m}\in \mathbb{R}^{n}$ 
are mapped into the $6+n$ dimensional space of generalized velocities by the transpose of actuator-selection matrix $\vS=\begin{bmatrix}
\text{\textit{\textbf{0}}}_{n \times 6} & \vI_{{n \times n}} \end{bmatrix} $. 
In this work, we divide the external torques acting on the robot into two components. The first one is assumed to be directly measured using a F/T sensor and is denoted $\vtau_{\sub{ft}}$. The second one is denoted as $\vtau_\sub{ext}$ and is due to external contact and collision forces that are not measurable with a F/T sensor, and disturbances. With a slight abuse of notation, in this paper we use the term \textit{disturbances} to refer to modeling inaccuracies, unmodeled payloads, sensor noise and errors in the estimation of motor torques $\vtau_{\sub{m}}$. The latter might occur when estimating the torque $\vtau_\sub{m}$ from current measurement (e.g., in pseudo direct drives) without any additional torque sensors.

\subsection{Torque estimation}
\label{sec:torque_estimation}
To estimate the external torques acting on the robot joints, $\vtau_\sub{ext}$, we use a momentum-based observer approach \cite{DeLuca2006CollisionArm}, which is based on the definition of the robot generalized momentum $\vp = \vM(\vq) \vv$.
Let $\vn(\vq, \vv)$ 
$:=\vg(\vq)+\vC(\vq, \vv) \vv-\dot{\vM}(\vq, \vv) \vv$.
Based on \cite{DeLuca2006CollisionArm}, $\hat \vtau_\sub{ext}$ can be computed as
\begin{equation} 
\label{eq:momentum_observer_continuous}
\hat{\vtau}_\sub{ext} = \vK_\sub{O}\left(\vp(t)-\int_{0}^{t}\vS^{T} \vtau_\sub{m} + \vtau_\sub{ft} -\hat{\vn}(\vq, \vv) +\hat{\vtau}_\sub{ext} \mathrm{d}s \right) ,
\end{equation}
where $\vK_\sub{O}$ is a positive diagonal observer gain matrix, and $\hat \vn$ is the estimate of the nonlinear terms $\vn$, obtained from the rigid body dynamics equations of the robot. 


\subsection{Wrench estimation}
\label{sec:wrench_estimation}
Let $\bm{\mathcal{F}}_{\sub{ext},i} = (\vF_{\sub{ext},i}, \vm_{\sub{ext},i}) \in \mathbb{R}^6$ be the wrench applied to the colliding link $i$. The force $\vF_{\sub{ext},i}$ is given by  $\vF_{\sub{ext},i} = \vF_{\sub{c},i} +  \vF_{\sub{dis},i}$, where $\vF_{\sub{c},i}$ is the external collision force, and $\vF_{\sub{dis}, i}$ is due to all other disturbance sources. Note that besides detecting the beginning and the end of a collision, the final goal of this paper is to isolate the disturbance component $\vF_{\sub{dis},i}$ and obtain an accurate estimate of the collision force $\vF_{\sub{c},i}$.

Once $\hat \vtau_\sub{ext}$ is obtained from Eq. \eqref{eq:momentum_observer_continuous}, the external forces and wrench are estimated as
\begin{equation} \label{eq:contact_wrench}
 \begin{bmatrix} \hat{\vF}_{\sub{f},1} \\ \vdots \\ \hat{\vF}_{\sub{f},4} \\ \hat{\bm{\mathcal{F}}}_{\sub{ext},i} \end{bmatrix} = \begin{bmatrix} \vJ_{\sub{f},1}^{T}(\vq) & \hdots & \vJ_{\sub{f},4}^{T}(\vq) & \vJ_{i}^{T}(\vq) \end{bmatrix}^{\#} \hat{\vtau}_\sub{ext} ,
\end{equation}
where the symbol $^{\#}$ denotes the Moore-Penrose pseudoinverse operation, $\vF_{\sub{f},j}$ is the force on contact foot $j$ with $\vJ_{\sub{f},j}\in \mathbb{R}^{3 \times (6+n)}$ its translational Jacobian, and $\vJ_{i}\in \mathbb{R}^{6 \times (6+n)}$ is the spatial Jacobian of the colliding link $i$ ~\cite{Vorndamme2017CollisionHumanoids, Flacco2016Residual-basedRobots}.
Here, the contact feet and the collision links are modeled as point contacts and thus are only subject to linear forces. However, since the exact contact location of the collision force is unknown, the spatial Jacobian of an arbitrary robot link is used in Eq~\eqref{eq:contact_wrench}, which results in the additional torque $\vm_{\sub{ext},i}$ caused by the impact force. 

In this work, we assume quasi-static scenarios. This assumption is valid even while trotting, since the arm and the base - on which the collisions occur - are not subject to high accelerations. 

\section{Collision detection, isolation and identification} \label{sec:collision_isolation_identification}
 
\begin{figure*}[t!]
\centering
\includegraphics[width=\textwidth]{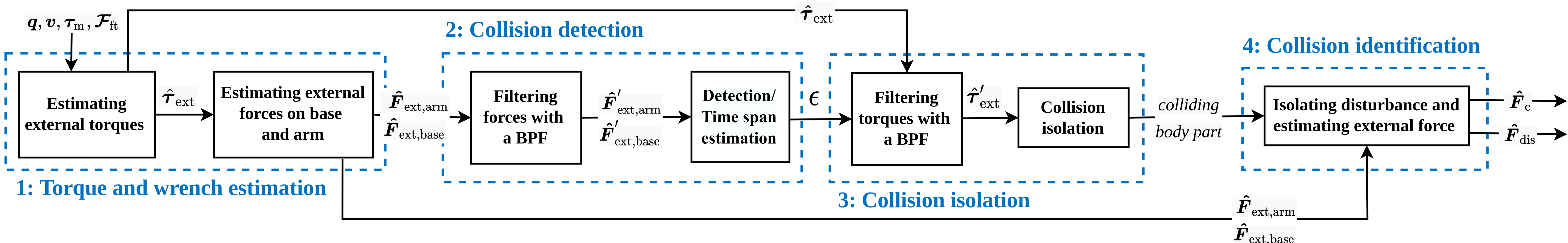}
\caption{Collision-event pipeline presented in this work. The pipeline consists of the following phases: torque and wrench estimation (Sec.~\ref{sec:torque_estimation}, \ref{sec:wrench_estimation}), collision detection (Sec.~\ref{sec:collision_detection}), isolation (Sec.~\ref{sec:collision_isolation_identification_sec:isolation}), and identification (Sec.~\ref{sec:collision_isolation_identification_sec:offset}).}
\label{fig:collision_pipeline}
\end{figure*}

As the fundamentals of torque and wrench estimation are explained in Sec.\ref{sec:preliminaries}, in this section we describe the remaining components of the pipeline: detection, isolation, and identification. A scheme of the pipeline is illustrated in Fig.~\ref{fig:collision_pipeline} and can be summarized in: (1) estimating the external torque $\hat \vtau_\sub{ext}$ and the external force at the base and arm; (2) filtering the estimated forces with a BPF to detect the start of a collision and estimate its time span; (3) filtering $\hat \vtau_\sub{ext}$ with a BPF and using it to determine if the arm or base is the colliding body part; and (4) isolating the disturbance from the force estimate to identify the collision force. With the final estimated collision force $\hat{\vF}_{\sub{c}}$, a reaction strategy can be implemented.

\subsection{Collision detection} 
\label{sec:collision_detection}


At the start of the collision pipeline, we do not have any knowledge of the colliding body part yet. Therefore, both the estimated external forces on the arm, $\hat \vF_\sub{ext,arm}$, and the base, $\hat \vF_\sub{ext,base}$, are used as inputs to the collision detection block (Fig.~\ref{fig:collision_pipeline}). They are estimated from Eq.~(\ref{eq:contact_wrench}) by stacking either the arm or the base Jacobian in $\vJ_i^T(\vq)$. While the estimated external torque vector $\hat \vtau_{\sub{ext}}$ could also be used for collision detection, it is more intuitive to use the estimated external forces at the arm and the base for observation and tuning purposes. We note that it is essential to verify both base and arm force for collision detection since particular arm collisions will not be detected in the base force, and vice versa. 

\subsubsection{Force filtering} During collision detection, we first filter $\hat \vF_{\sub{ext},\sub{arm}}$ and $\hat \vF_{\sub{ext},\sub{base}}$ with a BPF. We indicate the resulting filtered forces with $\hat \vF'_{\sub{ext},\sub{arm}}$ and $\hat \vF'_{\sub{ext},\sub{base}}$. While it has been proven that the use of a high-order BPF can increase the overall robustness of the detection ~\cite{Cho2013CollisionUncertainty,Cho2013Adaptation-and-collisionInteraction}, it increases the time delay compared to a first-order BPF~\cite{Manal2007AModeling}. We therefore opt for the latter in our experiments.

\subsubsection{Collision detection and time span estimation} 
\begin{figure}[!t]
\centerline{\includegraphics[width=\linewidth]{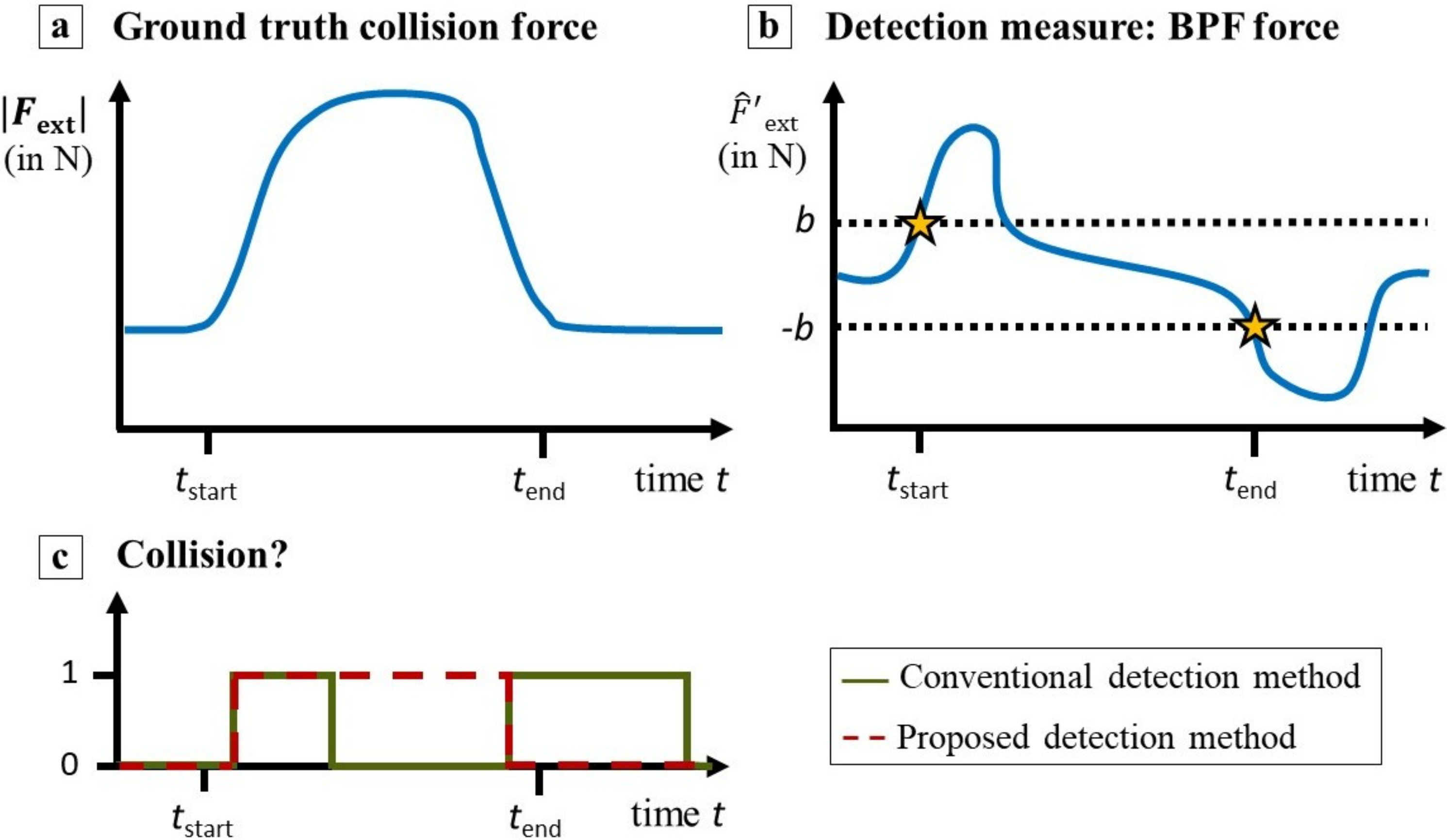}}
\caption{The collision force (\ref{fig:two_peaks_phenomenon}a) is estimated and filtered with a BPF (\ref{fig:two_peaks_phenomenon}b), where a constant threshold with value $b$ is set for detection and the stars indicate the detected start and end of the collision with the proposed method. The conventional detection approach works the same, but detects a collision only when the BPF force is exceeding the threshold. Thus, it would detect two collisions, while the proposed detection technique detects the time span accurately (\ref{fig:two_peaks_phenomenon}c).}
\label{fig:two_peaks_phenomenon}
\end{figure}

During a collision, the high-pass property of the BPF causes two peaks on the filtered forces, one at the start and one at the end of the impact. These peaks are the result of the high frequencies due to the sudden force change when the contact is applied and removed. This phenomenon is depicted Fig.~\ref{fig:two_peaks_phenomenon} (top-right plot). 
In the proposed method, we employ this two-peak BPF phenomenon 
to detect not only the beginning but also the end of the collision. 

Briefly, a collision is detected when the band-pass filtered force crosses the chosen threshold $b$ on either the positive or the negative half-plane. The end of the collision is then detected at the beginning of the second peak, i.e. when the signal crosses the threshold on the opposite half. This criterion is applied to each component of the band-pass filtered force. The detection of the beginning and the end of the collision using the band-pass filtered force, rather than the estimate of the force itself, is of particular importance as it is robust against disturbances. It also allows for an accurate estimation of the time span of the collision. 

The output of the collision detection is a variable $\epsilon$, which is 1 if a collision is detected and 0 otherwise.


\subsection{Collision isolation} \label{sec:collision_isolation_identification_sec:isolation}
A common approach to isolate the colliding body link is based on the assumption that, when a collision occurs at contact link $i$ in an open kinematic chain structure, this contact does not produce torques along the joints more distal from the link in contact \cite{Haddadin2017RobotIdentification, Vorndamme2017CollisionHumanoids, Ito2019ExperimentalRobot}. Consequently,  
the first link $i$ for which $|\hat{\tau}_{\sub{ext},i}| > b_{\text{is}}$, with $b_{\text{is}}$ the isolation threshold larger than 0, is defined as the one on which the external force is applied. The validity of this assumption can, however, be challenged as collision forces parallel to joint axes cannot be detected.


Here we aim to distinguish between base and arm collisions, hence retrieving the contact link and the exact contact location is out of the scope of this work.
Therefore, we use the following collision isolation rule: after a collision is detected, if one of the arm torques $\hat{\tau}'_{{ext},i}$ crosses its threshold, we conclude that the collision is occurring at the arm. Otherwise, it is occurring at the base.

\subsection{Collision identification} \label{sec:collision_isolation_identification_sec:offset}
As pointed out in Sec.\ref{sec:collision_detection}, applying a BPF to the estimated external force is a solution to make the collision detection algorithm robust to model inaccuracies and unmodeled loads.  However, this may alter the magnitude characteristics of the estimated force. Thus, the band-pass filtered forces cannot be used to identify the magnitude of the collision force.

Existing methods try to improve the force estimation accuracy by performing off-line model identification \cite{Mamedov2020PracticalDetection,Lim2021MomentumLearning}. However, this may be a complex and time-consuming process and errors will always remain \cite{Mamedov2020PracticalDetection}. Moreover, the identified model remains sensitive to changes in the robot configuration, wear and tear or accidents that might occur to the robot in a real-world environment.


Therefore, here we propose the following collision identification method.
For brevity, we define:
\begin{equation}
    \hat \vF_\sub{ext} =\begin{cases}
    \hat \vF_\sub{ext,arm}, \quad &\text{if arm collision} \\
    \hat \vF_\sub{ext,base}, \quad &\text{if base collision}
    \end{cases}
\end{equation}
as defined in Eq.~(\ref{eq:contact_wrench}). Let $\vF_\sub{dis}$ be the disturbance signal that includes modeling errors, unmodeled payloads and sensor noise. At time $k$, starting with $k=1$, we compute:
\begin{equation}
\label{eq:disturbance_signal}
    \hat \vF_\sub{dis}(k) =\begin{cases}
    \left(1-\alpha \right) \hat \vF_\sub{dis}(k-1) + \alpha \hat \vF_{\sub{ext}}(k), \; &\text{if } \epsilon = 0 \\
     \hat \vF_\sub{dis}(k-1), \; &\text{if } \epsilon = 1
    \end{cases}
\end{equation}
where $\alpha = e^{-\omega T_{{s}}}$, $\omega$ is the cut-off frequency of the low-pass filter (LPF) and $T_{{s}}$ is the sampling time. We assume no collision is occurring at $k=0$ and that $\hat \vF_\sub{dis}(0) =  \hat \vF_{\sub{ext}}(0)$. As shown in Eq.~\eqref{eq:disturbance_signal}, we assume that modeling errors stay constant while in collision ($\epsilon=1$), and thus we freeze $\hat \vF_{\sub{dis}}$ to its pre-collision value during this time. At time $k$, the collision force $\hat \vF_{\sub{c}}(k)$ is computed by subtracting the disturbance signal from the estimated value, i.e.: $ \hat \vF_{\sub{c}}(k) =  \hat \vF_{\sub{ext}}(k) - \hat \vF_\sub{dis}(k)$. 
\section{Experiments} \label{sec:experiments}
We verify the proposed pipeline through hardware experiments on the quadrupedal robot ANYmal with a 6DoF arm mounted on top \cite{Sleiman2021AManipulation} (Fig.~\ref{fig:experiments}). The robot is equipped with a RobotiQ 2F-85 gripper \footnote{https://robotiq.com/products/2f85-140-adaptive-robot-gripper} and a BOTA Rokubi SensOne 6-DoF F/T sensor \footnote{https://www.botasys.com/} at its end-effector. ANYmal’s legs are equipped with series elastic actuators while its arm is equipped with pseudo direct drives.



In total, in our experiments, we apply 416 collisions to the robot in various case studies visualized in Fig.~\ref{fig:experiments}. The collision distribution is presented in Table \ref{tab:datasets_collisions}. Examples of collisions in these scenarios are shown in the accompanying video \footnote{Available at \url{https://youtu.be/Rr4h4tEHCFU}}. Ground truth contact data is read from a hand-held F/T sensor; the collisions are created by either pushing the F/T sensor on different parts of the arm and base of the robot or by holding it still while the arm collides with it.

In the following sections, we first provide some experimental analysis to explain the reasoning behind our choice of continuous-time MBO for torque estimation (Sec.~\ref{sec:mbo-comparison}) and constant thresholds for detection (Sec.~\ref{sec:threhsold-comparison}). Afterwards, we validate the proposed collision-event pipeline in hardware tests (Sec.~\ref{sec:results}). 

\begin{table}[!t]
\ra{1.3}
\centering
\caption{Collision distribution during experiments.}
\resizebox{0.8\linewidth}{!}{
\begin{tabular}{@{}lll@{}}
\toprule \toprule
Scenario                                                                                     & \textbf{}            &Number of  collisions                                                                                                                                                               \\ \midrule
\textbf{Stance} & \textbf{no load} & 109 \\
\textbf{\begin{tabular}[c]{@{}l@{}} \end{tabular}}            & \textbf{measured load}                           & 133    \\
\textbf{\begin{tabular}[c]{@{}l@{}} \end{tabular}}            & \textbf{unmodeled load}                           & 99    \\                \hline                                                       
\textbf{Arm motion} & \textbf{no load} & 40 \\
\textbf{\begin{tabular}[c]{@{}l@{}} \end{tabular}}            & \textbf{measured load}                           & 10    \\
\textbf{\begin{tabular}[c]{@{}l@{}} \end{tabular}}            & \textbf{unmodeled load}                           & 13      
\\ \hline
\textbf{Trotting}& \textbf{no load} & 12\\
\bottomrule \bottomrule
\end{tabular}
}   
\label{tab:datasets_collisions}
\end{table}

\subsection{Comparison of torque estimation methods}
\label{sec:mbo-comparison}
\begin{figure}[!t]
\centerline{\includegraphics[width=\linewidth ]{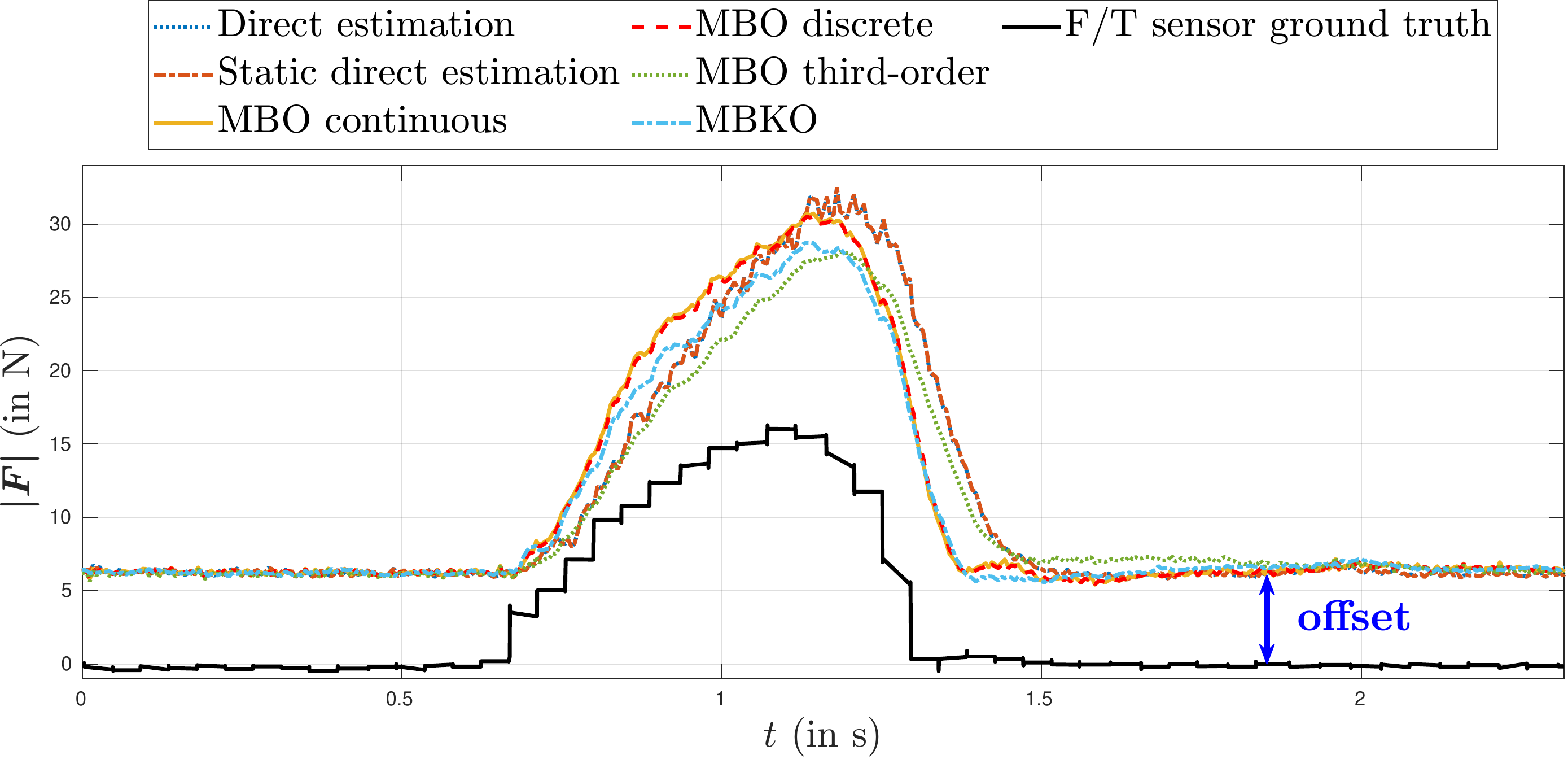}}
\caption{One of the collisions applied on the arm of the robot during stance. Due to disturbances, an offset is visible between the estimated and the ground truth force.}
\label{fig:compare_observers}
\end{figure}

In this section, we compare existing variants of MBOs in hardware experiments. The purpose of this comparison is to decide which torque estimation method is most suitable to be used with our collision identification approach, which are evaluated separately in Sec.~\ref{sec:results}. We consider scenarios where collisions are applied on the arm, while the robot is in stance (Fig.~\ref{fig:compare_observers}). We consider the following methods:
\begin{itemize}
    \item \textbf{Direct estimation} \cite{Ito2019ExperimentalRobot}: computes $\hat{\vtau}_{\sub{ext}}$ from Eq.~\eqref{eq:dynamic_model_floating_base}.
    \item \textbf{Static direct estimation} \cite{Mattioli2017InteractionRobots}: similar to \cite{Ito2019ExperimentalRobot}, but neglects the acceleration and velocity-dependent terms.
    \item \textbf{Continuous-time MBO} \cite{Vorndamme2017CollisionHumanoids}: implements Eq.~\eqref{eq:momentum_observer_continuous}.
    \item \textbf{Discrete-time MBO} \cite{Bledt2018ContactTerrains}: discretized implementation of the MBO
    \cite{Vorndamme2017CollisionHumanoids};
    \item \textbf{Third-order MBO} \cite{Morlando2021Whole-bodyRobots}: similar to the continuous-time MBO, but using a higher-order LPF. This results in more accurate estimations due to a sharper filtering action.
    \item \textbf{MBKO} \cite{Wahrburg2015CartesianMomentum}: reformulates the continuous-time MBO \cite{Vorndamme2017CollisionHumanoids} as a Kalman filter.
\end{itemize}
Except for the MBKO \cite{Wahrburg2015CartesianMomentum}, all these approaches have been used on legged robots in previous literature. A comparison was therefore essential to understand the differences between them and their impact in the scenario at hand.


\begin{table}[t!]
\ra{1.3}
\centering
\caption{Comparison of torque estimation methods. 
}
\resizebox{\columnwidth}{!}{
\begin{tabular}{@{}llllll@{}}
\toprule \toprule
                           & \begin{tabular}[c]{@{}l@{}}Tuning \\ simplicity\end{tabular} & \begin{tabular}[c]{@{}l@{}}Computation\\ time\end{tabular} & \begin{tabular}[c]{@{}l@{}}Noise \\ (in N)\end{tabular} & \begin{tabular}[c]{@{}l@{}}Delay \\ (in ms)\end{tabular} & \begin{tabular}[c]{@{}l@{}}Absolute \\ error\\ (in \%)\end{tabular} \\ \midrule
\textbf{Direct estimation} &    {\color[HTML]{009901} $\bm{++}$ }                                                          &                                            $+$                 & {\color[HTML]{FE0000} \textbf{0.72}}                                              & 179                                                                 & {\color[HTML]{FE0000} \textbf{37 }}                          \\
\textbf{Static direct estimation}       &     {\color[HTML]{009901} $\bm{++}$ }                                                            &                                           $+$                   & {\color[HTML]{FE0000} \textbf{0.72}}                                              & 180                                                                   & {\color[HTML]{FE0000} \textbf{37}}                            \\
\textbf{MBO continuous}    &     $+$                                                          &                                            $+$               & 0.30                                                                              & {\color[HTML]{009901} \textbf{159}}                                                                                                        & 34                                                            \\
\textbf{MBO discrete}      &   $+$                                                           &                                                  $+$            & 0.30                                                                              & {\color[HTML]{009901} \textbf{160}}                                                                                                        & 33                                                             \\
\textbf{MBO third-order}   &   {\color[HTML]{FE0000} $\bm{-}$}                                                            &                                                    $+$          & 0.32                                                                              & {\color[HTML]{FE0000} \textbf{185}}                                                                    & {\color[HTML]{009901} \textbf{31}}                            \\
\textbf{MBKO}              &      {\color[HTML]{FE0000} $\bm{-}$}                                                          &                                                   {\color[HTML]{FE0000} $\bm{--}$}           & {\color[HTML]{009901} \textbf{0.21}}                                              & 176                                                                                                    &{\color[HTML]{009901} \textbf{30}}                                                               \\ \bottomrule \bottomrule
\end{tabular}
}
\label{tab:compare_observers}
\end{table}

We use the methods above to obtain an estimate of the generalized external torque $\vtau_\sub{ext}$. Afterwards, we compute the estimated external force $\hat \vF_{\sub{ext,arm}}$ as explained in Sec.\ref{sec:wrench_estimation}. The results from our comparison, considering 40 collisions in stance and arm motion, are reported in Table \ref{tab:compare_observers}. In Fig.~\ref{fig:compare_observers}, the response of the selected methods to one of the collisions applied to the arm of the robot is visualized. In Table \ref{tab:compare_observers}, we use \textit{noise} to refer to the standard deviation over the time in between collisions. The \textit{delay} is defined as the detection time. To compute this detection time, we set a constant threshold on the band-pass filtered forces, similar to the threshold set at $10\%$ of the estimated force in \cite{Birjandi2020Observer-extendedSensing}. This threshold is set equal for each method. The \textit{absolute error} is defined as:
\begin{equation} \label{eq:absolute_error}
    e = \left|\left( \frac{\left|\hat \vF^{*}\right|}{\left| \vF^{*}\right|}-1\right)\cdot 100 \% \right| , 
\end{equation}
where $|\hat \vF^{*}|$ and $|\vF^{*}|$ indicate the magnitude peak values of the estimated and ground-truth F/T sensor collision force, respectively. Note that we compare the methods based on the estimated contact force magnitude and we do thus not consider force direction.

As shown in Fig.~\ref{fig:compare_observers}, during the experiments the disturbance effect is not negligible and is reflected in a constant offset of about 7 N.
Thus, when computing the absolute error in Table~\ref{tab:compare_observers}, we take into account the offset for all the methods to obtain a fair comparison.

It is visible from Table~\ref{tab:compare_observers} that the methods based on the direct estimation of $\vtau_\sub{ext}$
have a high absolute error. In addition, the force estimated with such methods shows oscillations when the force estimate is at its peak value. Direct estimation shows higher delay than the MBO because it relies on accelerations which are quite noisy especially, for example, in the trotting case and need to be heavily filtered.
The third-order MBO and MBKO have the highest estimation accuracy in terms of absolute error and noise, respectively. However, computation time is significantly higher for the MBKO, and the third-order MBO has a large delay. Furthermore, both methods have a higher tuning complexity, which is an important aspect to consider when working with a high-DoF quadrupedal manipulator. The higher-order filter could also possibly result in oscillations and instability \cite{Priemer2018FundamentalsProcessing}.
Thus, we conclude that the continuous-time and discrete-time MBOs provide the best trade-off between delay and estimation accuracy. Since they show similar results, we select the continuous-time MBO for our experiments.

\subsection{Threshold comparison}
\label{sec:threhsold-comparison}
We compare constant and dynamic thresholds to assess how they influence the collision detection performance. We consider a dynamic threshold based on joint velocities from \cite{Geravand2013Human-robotArchitecture} and a dynamic threshold based on estimated force standard deviation from \cite{Escobedo2021ContactSensors} (although we do not include proximity sensors). 
\begin{figure}[!t]
\centerline{\includegraphics[width=\linewidth]{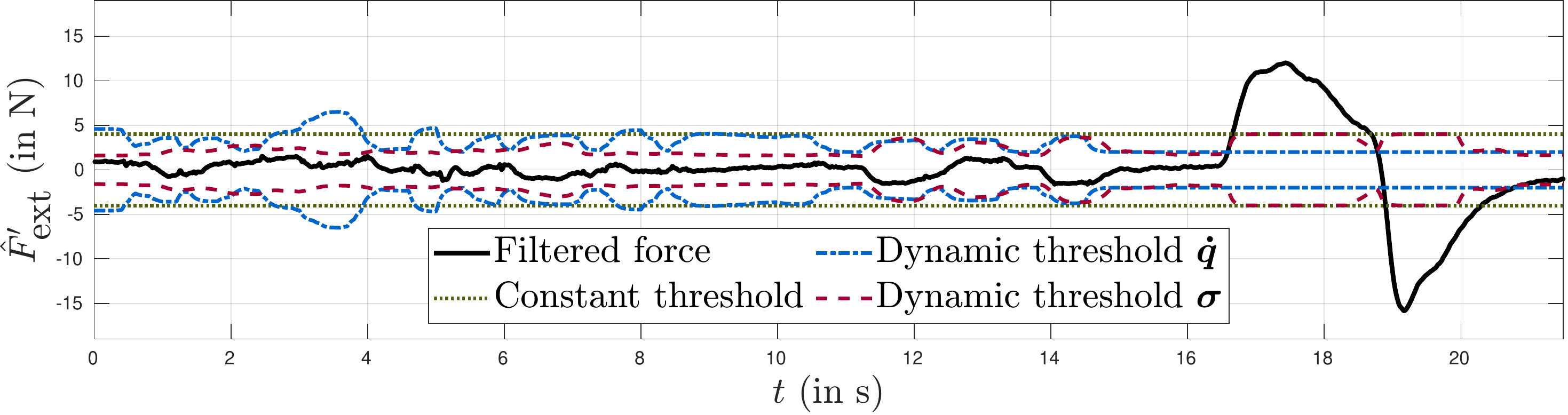}}
\caption{Comparison of constant and dynamic thresholds. The dynamic threshold is based on the desired joint velocity $\dot\vq_\sub{d}$ and on estimated force standard deviation $\vsigma$. The arm is moving until a collision occurs at $t = \unit[16.5]{s}$.
}
\label{fig:dynamic_thresholds}
\end{figure}

An average of $\unit[14]{ms}$ improvement in detection delay is obtained using a dynamic threshold; this result was obtained in experiments that we conducted with the robot in stance, performing arm motion, and in the presence of an unmodeled payload. To tune the constant threshold and the constant parts of the dynamic threshold equations, we evaluate all collisions during arm motion. Thresholds are set such that no FPs and FNs occur and are set to the lowest values possible. Since we consider arm motion, which makes the force estimation noisy, the tuned thresholds should be robust against FPs, meaning there is no need to adjust these in other scenarios. A plot from one of the experiments is reported in Fig.~\ref{fig:dynamic_thresholds}. 
For conciseness, in Fig.~\ref{fig:dynamic_thresholds}, we only plot the component of the estimated force along the main contact direction.

The dynamic thresholds move along with the variations in the filtered force that arise due to model errors. Thus, they are helpful to increase detection robustness in cases such as trotting or arm motion compared to when the robot is in a static configuration. A static threshold needs to be conservative to avoid FPs resulting from such scenarios where $\hat \vtau_\sub{ext}$ may increase due to high accelerations. 
This is especially the case for our experiments which include trotting, where such variations can arise due to the high-frequency impact of the robot's feet on the ground. 
In this work, we opted for using three different static thresholds for the following scenarios: trotting, arm motion and stance. This proved sufficiently robust and did not exhibit any decrease in performance w.r.t. dynamic thresholds which, on the other hand, remain sensitive to tuning parameters.

\subsection{Collision detection and identification results}
\label{sec:results}
 In this section, we present the experimental results of our pipeline and discuss our findings. We test our collision identification method in different scenarios in the presence of the following factors of variation: 
\begin{itemize}
    \item \textbf{Mode.} Stance, arm motion and trotting.
    \item \textbf{External load.} As shown in Fig.\ref{fig:experiments}, we add a payload on different links of the robot. In particular, we consider the following cases:\begin{itemize} \item \textit{Unmodeled load}: an unmodeled $\unit[0.58]{kg}$ payload is placed on arm or base, or a $\unit[2]{kg}$ load on the base; \item \textit{Measured load}: a load of $\unit[0.58]{kg}$ is added to the gripper with its force measured by the F/T sensor placed at the robot end-effector. \end{itemize}
    \item \textbf{Duration and magnitude of force.} Magnitudes of the applied force range in the interval $\unit[5-165]{N}$, and time span of collision is in $\unit[0.3-6.0]{s}$. 
\end{itemize} 
In all the experiments we use a cut-off frequency $f = \unit[0.5]{Hz}$, with $\omega = 2 \cdot \pi \cdot f$, for estimating the disturbance force $\vF_\sub{dis}$ of Eq.~\eqref{eq:disturbance_signal}. The cut-off frequencies of the BPF for $\hat \vF_{\sub{ext}}$ are selected as $\unit[0.4, 3.0]{Hz}$. 

\subsubsection{Collision detection and time span estimation}
We validate our collision detection algorithm over a dataset of 416 collision experiments on both arm and base. Note that we assume detection of fast-rising contact forces only. We define the \textit{success rate} and the \textit{precision} over $N$ collisions as $(N-N_\sub{FN})/N$ and $N/(N+N_\sub{FP})$, respectively, where $N_\sub{FN}$ is the number of FNs, and $N_\sub{FP}$ is the number of FPs. Overall, we achieve a $\unit[99]{\%}$ success rate 
and a precision of $\unit[98]{\%}$ for collision detection. We can detect all collisions during stance and arm motion with and without unmodeled loads, i.e. all scenarios mentioned in Fig.~\ref{fig:experiments} except for trotting. All undetected collisions occur during trotting and represent very challenging scenarios, such as when the magnitude of the collision force is below $\unit[10]{N}$. Note that the noise of the raw MBO output during trotting can reach up to 35 N as can be seen in Fig.~\ref{fig:collision_motion_trot:trot}.

\begin{figure}[!t]
\centerline{\includegraphics[width=\linewidth]{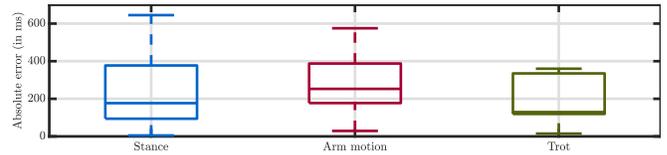}}
\caption{Boxplots comparing the absolute estimation error of the collision time span.}
\label{fig:boxplot_time_span}
\end{figure}

Next, in  Fig.~\ref{fig:boxplot_time_span} we analyze the time span estimation accuracy of our proposed approach over a set of 156 detected collisions performed in stance, arm motion and trotting without considering additional loads. The average absolute error of time span estimation is $\unit[234]{ms}$, $\unit[272]{ms}$ and $\unit[390]{ms}$, respectively, for the previously mentioned three scenarios.

\subsubsection{Robustness analysis of collision identification}
\begin{figure}[!t]
	\centering
	\begin{minipage}{\columnwidth}
	    \hspace{5mm}
		\includegraphics[width=0.93\textwidth]{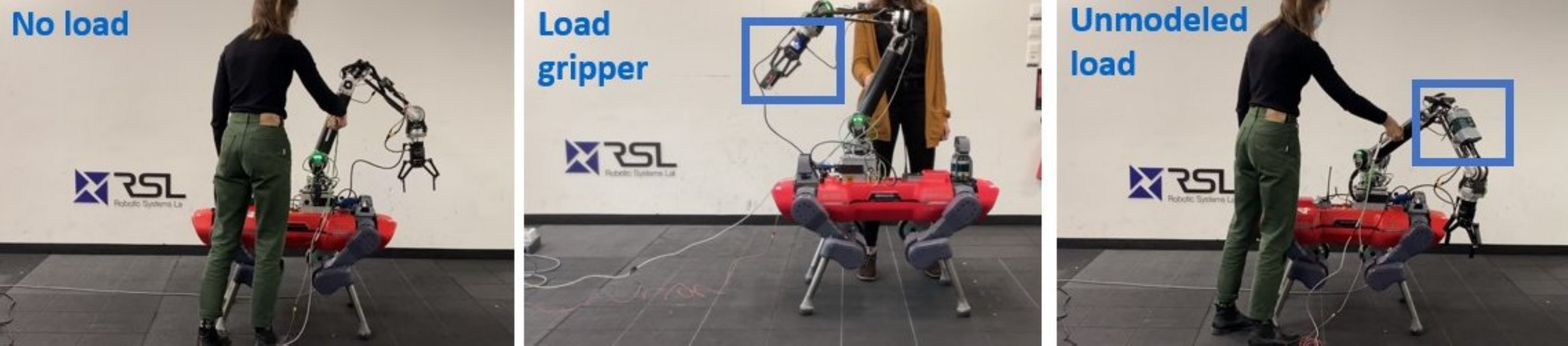}
	\end{minipage}%
	\hfill
	\vspace{3mm}
	\begin{minipage}{\columnwidth}
		\centering
		\includegraphics[width=\textwidth]{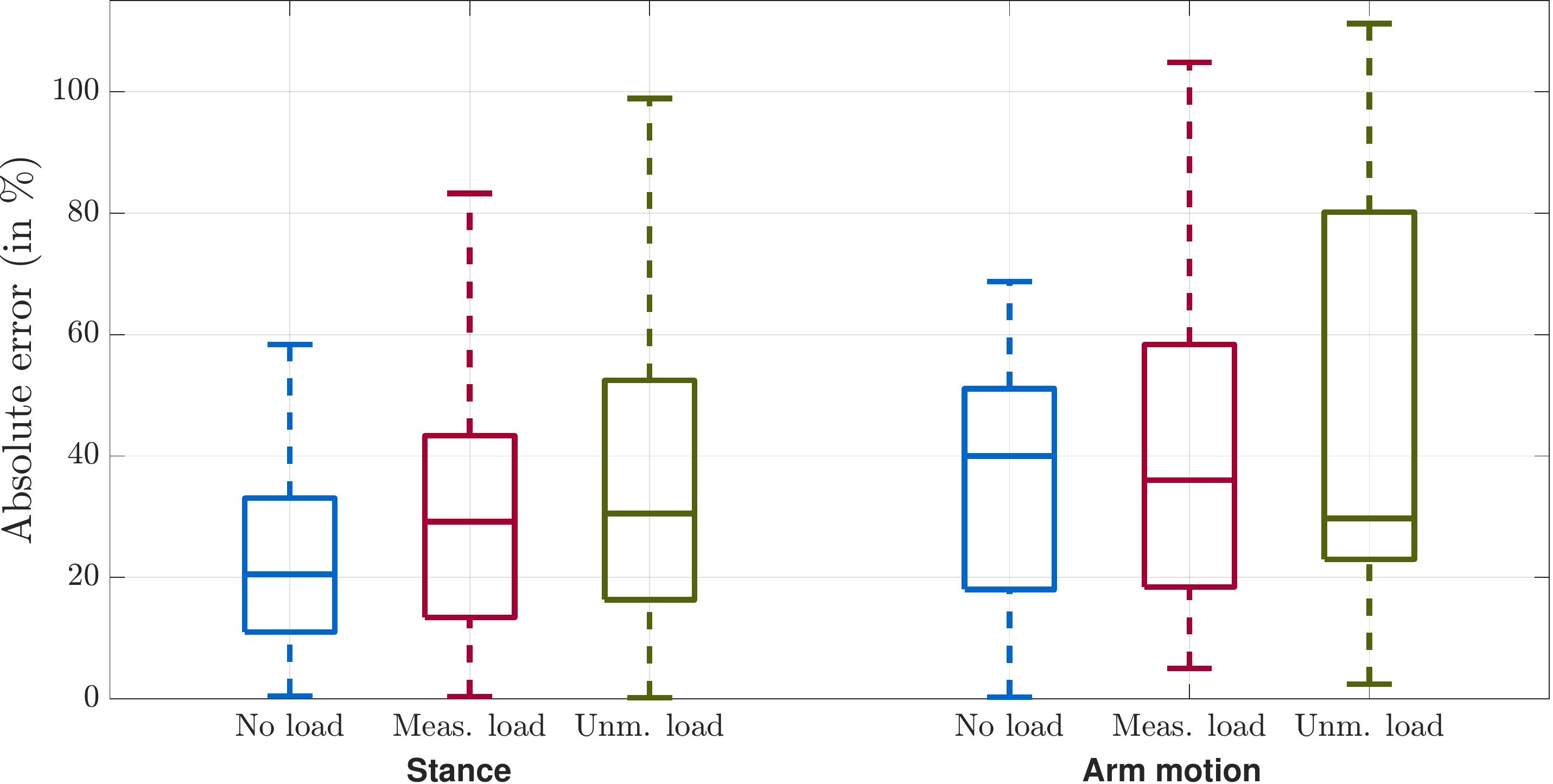}
	\end{minipage}
\caption{Boxplots comparing the absolute estimation error of the external force magnitude in various scenarios. `Meas. load' indicates presence of an object in the gripper and `Unm. load' an object on the arm or base.}
\label{fig:boxplot_scenario_comparison}
\end{figure}

We compare the performance of our method in three different cases: \textit{No load}, \textit{Measured load} and \textit{Unmodeled load} of different weights, placed on various locations on the robot.
The collision force estimation accuracy of the proposed identification method in these scenarios is visualized in Fig.~\ref{fig:boxplot_scenario_comparison}.
The variability is higher in the \textit{Measured load} and \textit{Unmodeled load} scenarios, compared to the \textit{No load} scenario. This is because of the increased inertia of the robot caused by the payload, which results in a more significant response to a collision. 
Additionally, the medians of the absolute errors between the three scenarios are comparable. This underlines the fact that our collision identification approach compensates robustly for the unmodeled load, due to the continuous estimation of the disturbance force $\vF_\sub{dis}$, as introduced in Sec.~\ref{sec:collision_isolation_identification_sec:offset}. 

Furthermore, we evaluate the repeatability of our collision identification method. To do this, we split a dataset of 73 arm and base collisions in stance into 25 sets of doubles, triples, and quadruples of contacts with equal magnitude and location. 
The average of the standard deviation of the error computed within these sets is $\unit[5]{\%}$, compared to a standard deviation of the error of $\unit[12]{\%}$ within all collisions without the split.


\subsubsection{Comparison against baseline collision identification methods}
To validate the performance of our collision identification approach, we compare it to the state-of-the-art methods from Sec.~\ref{sec:mbo-comparison} that have been verified on legged robots. Hence, we select the continuous-time MBO from \cite{Vorndamme2017CollisionHumanoids}, discrete-time MBO \cite{Bledt2018ContactTerrains} and third-order MBO \cite{Morlando2021Whole-bodyRobots}. 
As explained in Sec.~\ref{sec:mbo-comparison}, we use Eq.~\eqref{eq:contact_wrench} to obtain the estimated external forces $\hat \vF_{\sub{ext,arm}}$ or $\hat \vF_{\sub{ext,base}}$. 
We carry out two different studies. 

\begin{figure}[!t]
\centering
\begin{subfigure}[t]{\linewidth}
         \centering
         \includegraphics[width=\textwidth]{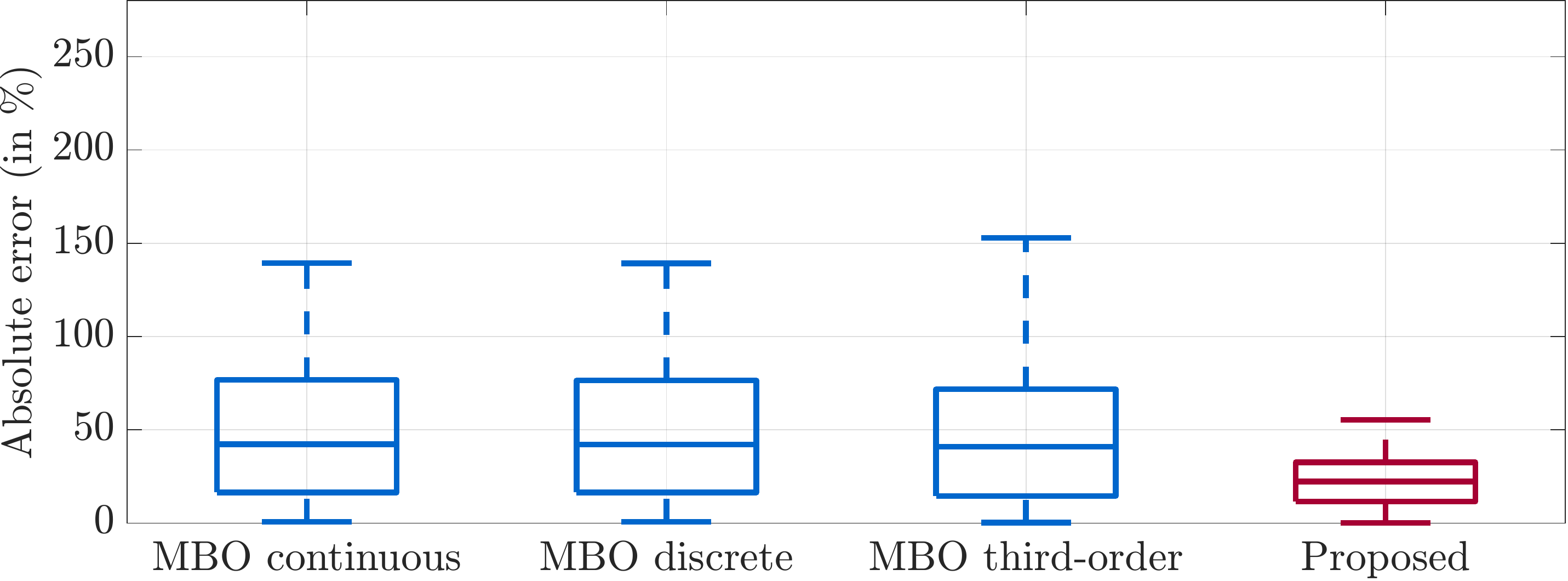}
         \caption{Collisions on base and arm during stance. }
     \end{subfigure}
     \hfill
     \vspace{1mm}
     \begin{subfigure}[t]{\linewidth}
         \centering
         \includegraphics[width=\textwidth]{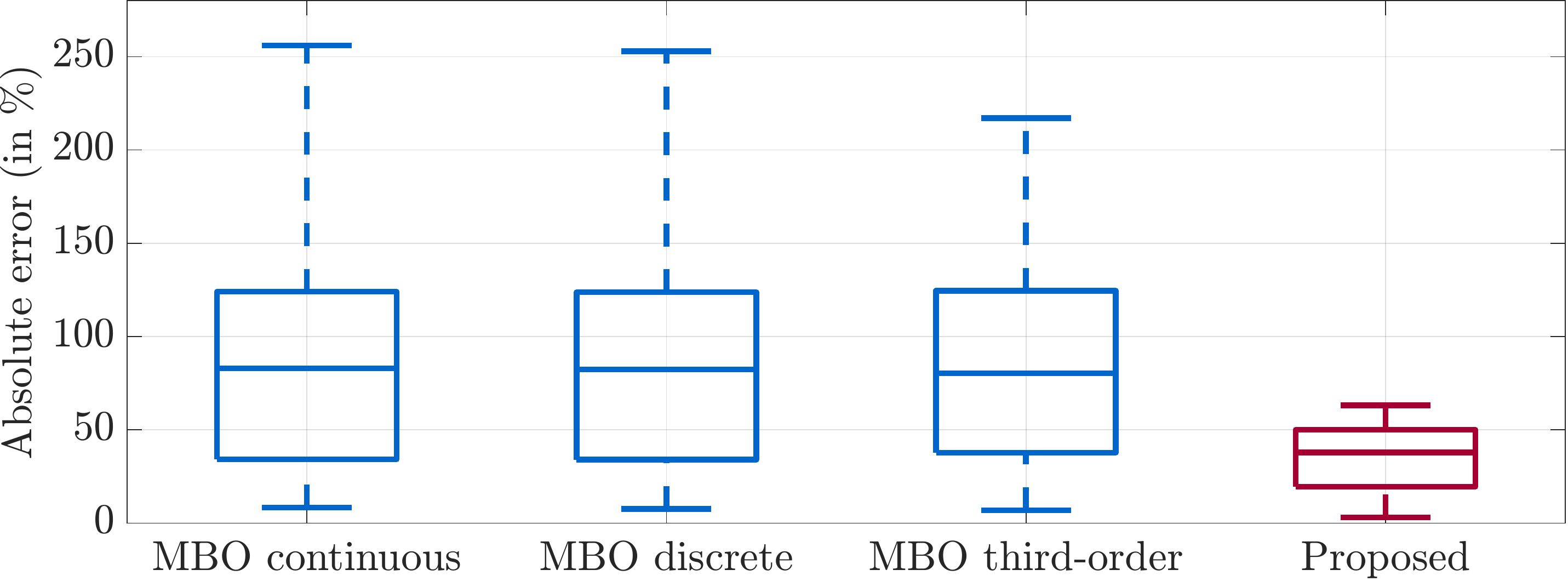}
         \caption{Collisions on base and arm during arm motion.}
     \end{subfigure}
     \hfill
     \vspace{1mm}
     \begin{subfigure}[t]{\linewidth}
         \centering
         \includegraphics[width=\textwidth]{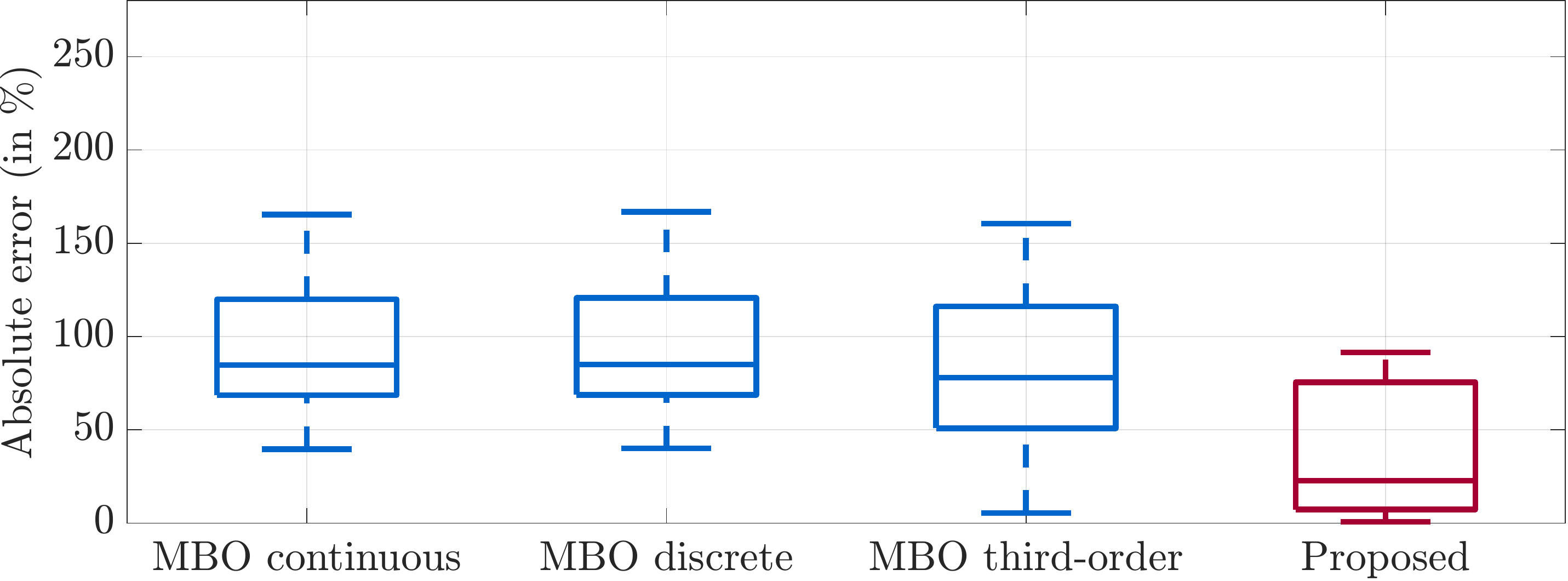}
         \caption{Collisions on arm during trotting.}
     \end{subfigure}
\caption{Boxplots comparing the external force magnitude estimation of the following collision identification methods: continous-time MBO, discrete-time MBO, third-order MBO introduced in Sec.~\ref{sec:mbo-comparison}, and ours.}
\label{fig:boxplot_offset_subtraction_comparison}
\end{figure}

First, we compare the overall performance of our method in three scenarios: stance, arm movement and trotting, without additional loads on the robot. We point out that our goal is to estimate the collision, and not the payload force. The results for the first comparison study are shown in Fig.~\ref{fig:boxplot_offset_subtraction_comparison}. The error metric $e$ from Eq.~\eqref{eq:absolute_error} is used to compute the error between the identified collision force $\hat \vF_\sub{c}$ and the ground-truth F/T sensor collision force.
With our method, a significant improvement in collision estimation accuracy can be achieved in various scenarios. The average absolute error is reduced from $\unit[52]{\%}$ in methods \cite{Vorndamme2017CollisionHumanoids, Bledt2018ContactTerrains} and $\unit[49]{\%}$ in \cite{Morlando2021Whole-bodyRobots} to $\unit[23]{\%}$ during stance. While for trotting, the average error is reduced from $\unit[100,101]{\%}$ in \cite{Vorndamme2017CollisionHumanoids, Bledt2018ContactTerrains} and $\unit[87]{\%}$ in \cite{Morlando2021Whole-bodyRobots} to $\unit[42]{\%}$. 
Note that due to the high-frequency high impact forces of the feet, in this case, the estimated forces show a large noise level. Therefore, we add an LPF to the estimated forces during trotting for all the methods in the comparison.

\begin{figure}[!t]
\centering
    \begin{subfigure}[t]{\linewidth}
         \centering
         \includegraphics[width=\textwidth]{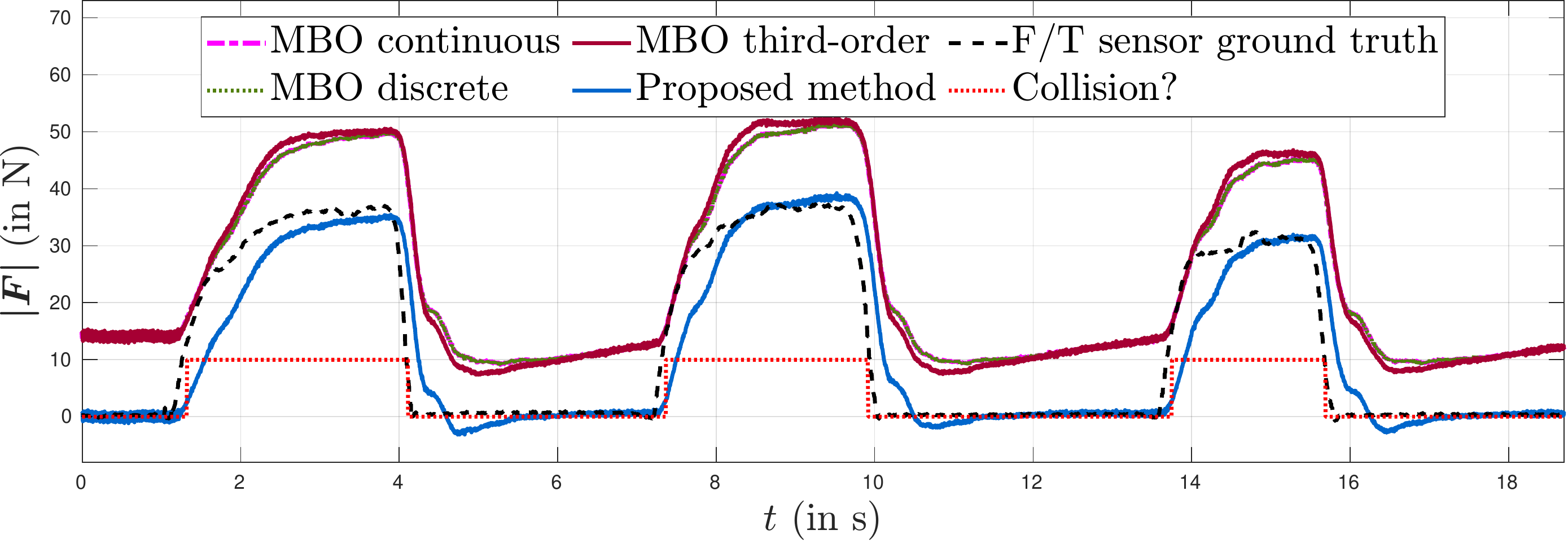}
         \caption{Three arm collisions with an unmodeled payload on the arm.}
         \label{fig:collision_motion_trot:load}
     \end{subfigure}
     \hfill
     \begin{subfigure}[t]{\linewidth}
         \centering
         \includegraphics[width=\textwidth]{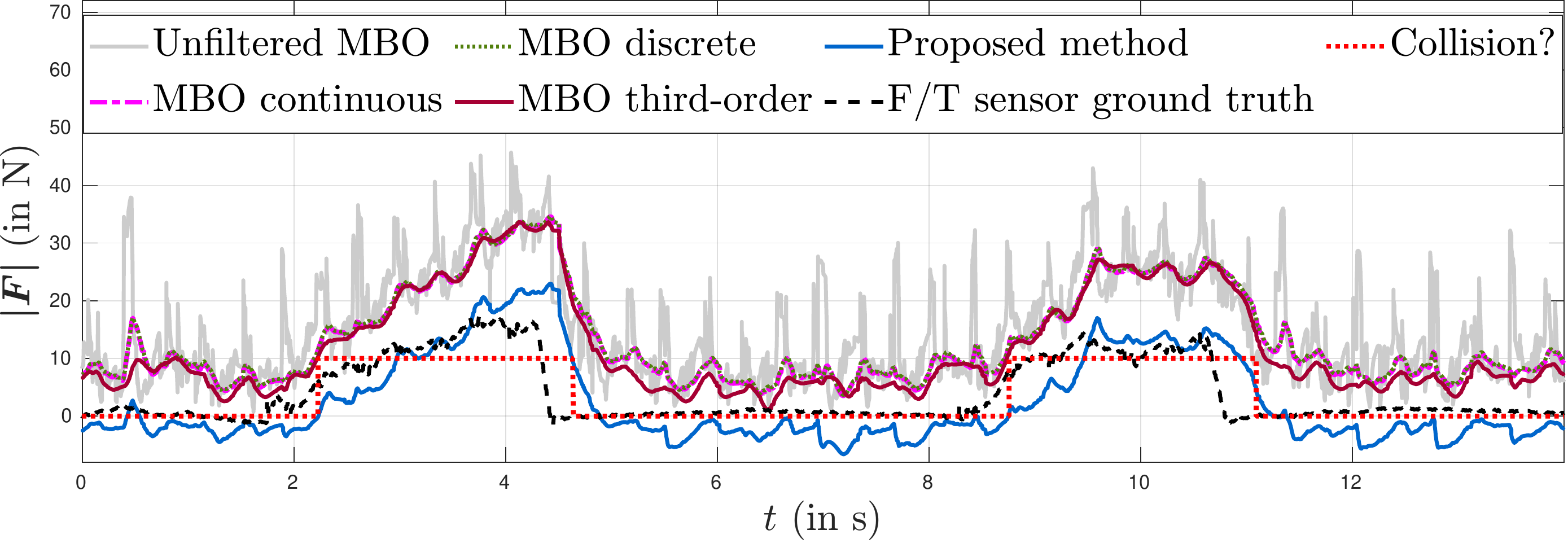}
         \caption{Two arm collisions during trotting. Note that an LPF has been added to the estimated forces.}
         \label{fig:collision_motion_trot:trot}
     \end{subfigure}
\caption{Comparison of external collision identification methods.}
\label{fig:collision_motion_trot}
\end{figure}

Secondly, we show a detailed analysis of the performance in two collision
scenarios: with an additional load on the arm and during trotting. The corresponding plots are presented in Fig.~\ref{fig:collision_motion_trot}. 
Note that the offset in the forces estimated with the conventional MBOs is due to errors coming from modeling inaccuracies and the unmodeled payload (in Fig.~\ref{fig:collision_motion_trot:load}).
As shown in Fig.~\ref{fig:collision_motion_trot:load}, state-of-the-art methods cannot compensate for the unmodeled payload, while our method is able to do so.

Moreover, trotting is a challenging scenario with many high frequency variations in the estimated forces, as it can be seen in the unfiltered MBO signal presented in Fig.~\ref{fig:collision_motion_trot:trot}. However, the trend line of the two collision forces is followed well and the absolute error is reduced with our approach: from $\unit[98, 97]{\%}$ \cite{Vorndamme2017CollisionHumanoids, Bledt2018ContactTerrains} and $\unit[92]{\%}$ \cite{Morlando2021Whole-bodyRobots} to $\unit[31]{\%}$ during the first collision, and from $\unit[102]{\%}$ \cite{Vorndamme2017CollisionHumanoids, Bledt2018ContactTerrains} and $\unit[88]{\%}$ \cite{Morlando2021Whole-bodyRobots} to $\unit[18]{\%}$ during the second. 

\section{Conclusions and future work} 
\label{sec:conclusion}

In this paper, we introduce a collision event pipeline for quadrupedal manipulators. This includes a method for computing the time span of collisions in the presence of model inaccuracies and unmodeled loads, and an improved identification of the collision force itself. 
We verify our approach by carrying out extensive hardware experiments, involving loads placed at different points on the robot and trotting. We produce a large dataset of collisions, that we make publicly available. A comparison with other state-of-the-art approaches is also presented. 

 As future work, we aim to additionally classify collisions and design appropriate reaction strategies. Moreover, to improve isolation, approaches that have been validated on manipulators, such as Bayesian filtering \cite{Bimbo2019CollisionSensing} 
 and machine learning \cite{Liang2021ContactSensing}
 could be extended to the legged robot case. 

\bibliographystyle{ieeetr}      
\bibliography{references_copy.bib}

\end{document}